\documentclass{article}
\usepackage{spconf,amsmath,graphicx}
\usepackage{amssymb, amsmath}
\usepackage{multirow}
\usepackage{graphics}
\usepackage{graphicx}
\usepackage{footnote}
\usepackage{hyperref}

\usepackage{enumitem}
\setlist{nosep, leftmargin=14pt}

\renewenvironment{thebibliography}[1]{
  \begin{oldthebibliography}{#1}
    \setlength{\itemsep}{0.5em}
    \setlength{\parskip}{0em}
}
{
  \end{oldthebibliography}
}

\usepackage{mwe} 

\title{Anomaly Detection in Retinal Images using Multi-Scale Deep Feature Sparse Coding}
\name{Sourya Dipta Das$^1$, \enspace Saikat Dutta$^2$, \enspace Nisarg A. Shah$^3$, \enspace Dwarikanath Mahapatra$^{4,5}$, \enspace Zongyuan Ge$^{5,6}$}
\address{$^1$Jadavpur University, India \enspace $^2$IIT Madras, India \enspace $^3$IIT Jodhpur, India \enspace $^4$Inception Institute of AI, Abu Dhabi \\ \enspace $^5$Monash University, Australia \enspace $^6$Airdoc-Monash Research, Australia}

\begin{document}
%
\maketitle
\begin{abstract}
Convolutional Neural Network models have successfully detected retinal illness from optical coherence tomography (OCT) and fundus images. These CNN models frequently rely on vast amounts of labeled data for training, difficult to obtain, especially for rare diseases. Furthermore, a deep learning system trained on a data set with only one or a few diseases cannot detect other diseases, limiting the system's practical use in disease identification. We have introduced an unsupervised approach for detecting anomalies in retinal images to overcome this issue. We have proposed a simple, memory efficient, easy to train method which followed a multi-step training technique that incorporated autoencoder training and Multi-Scale Deep Feature Sparse Coding (MDFSC), an extended version of normal sparse coding, to accommodate diverse types of retinal datasets. 
We achieve relative AUC score improvement of 7.8\%, 6.7\% and 12.1\% over state-of-the-art SPADE on Eye-Q, IDRiD and OCTID datasets respectively.

\end{abstract}
\vspace{-5px}
\section{Introduction}
\label{sec:intro}
Anomaly detection is a widely discussed topic in the Machine learning community. Ocular illnesses such as diabetic retinopathy (DR), age-related macular degeneration (AMD), Macular Hole, Central Serous Retinopathy, and glaucoma impact more than 270 million people globally. Anomaly detection in retinal data is a significant problem that is useful in identifying any abnormalities in the patients \cite{fernando2020deep}. Supervised classification algorithms \cite{gammulle2020two, schmidt2018artificial, esteva2017dermatologist} are often used to classify normal and anomalous data. However, training supervised classifiers require a good amount of annotated data which is often hard to obtain in the field of retinal imaging. Even when annotated data is available, the classifiers might suffer from the class imbalance as the prevalence of normal samples is frequently higher than that of abnormal samples. Also, some lesions are uncommon, and the presence of specific lesions is unknown until the diagnosis is made. Moreover, the labels collected from different clinical experts may be different, and hence the annotation process can produce noisy or biased labels \cite{deep_if}. 

In this context, anomaly detection in an unsupervised manner can be helpful. 

In our proposed method, we initially train an autoencoder to reconstruct image patches from normal images to learn domain-specific fine-grained features. 
This step will help learn more relevant micro and macro-level features unique to normal samples in local and global contexts.
Once the autoencoder is trained, we utilize multi-scale features extracted from the encoder's multiple layers from various depths as an input vector for sparse coding. 
With multi-scale deep features, it will capture more global context from the image to detect anomalies with different scales or sizes.
Our method has outperformed state-of-the-art unsupervised algorithms on three different retinal datasets. 

Our main contributions in this paper are
It is simple, memory efficient, and easy to train, unlike other methods,
We have proposed a multi-step training strategy that combines autoencoder training and Multi-Scale Deep Feature Sparse Coding (MDFSC) for anomaly detection to adopt a different type of datasets,
We extended Multi-Scale Deep Feature Sparse Coding (MDFSC) from sparse coding for various types of retinal datasets.

\vspace{-10px}
\section{Related work}
\vspace{-0.1cm}
\label{sec:rel_work}
In this section, we have discussed few previous state-of-the-art works related to retinal diseases. Zhou et al. \cite{zhou2020sparse} presented a Sparse-GAN constrained by a novel Sparsity Regularization Net to map input images into a latent space with a sparsity regularizer and estimate anomaly score in the latent space domain. However, they only designed and evaluated their method on OCT-based data. Cohen et al. \cite{SPADE} proposed Semantic Pyramid Anomaly Detection (SPADE) method based on correspondences between an anomalous image and a few similar normal images by using a multi-resolution feature pyramid. Ouardini et al. \cite{deep_if} presented an effective transfer learning method by using an Imagenet pre-trained Inception-ResNet-v2 model as a feature extractor and fitted an Isolation Forest model for anomaly detection as it can handle minimal tuning requirements. Schlegl et al. \cite{schlegl2017unsupervised} proposed a generative adversarial network-based model, AnoGAN, to learn discriminative features to encode images in latent space by accounting for normal anatomical variability.
Nevertheless, this network is hard to train and may get trapped into local minima. Golan et al. \cite{deep_geo} proposed a multi-class model to learn class-specific salient geometrical features by discriminating between several geometric transformations of normal images, and have used softmax activation statistics for anomaly detection. Imamura et al. \cite{imamura2021mlf} presented a sparse-coding-based method that used activation maps from different layers of Imagenet pretrained image classifier as input feature vectors. But for out-of-domain datasets, it performs poorly. Zhou et al. \cite{zhou2020encoding} proposed a P-net model which used two sub-networks: structure extraction network and image reconstruction network to extract structure and texture-based features for anomaly detection. However, their structure extraction network needs to train on a different dataset (e.g., segmentation, detection, etc.) with the similar domain.
On the contrary, our proposed method is simple, robust, efficient, easy to train that learns a dictionary in sparse coding using multi-scale deep features extracted from a trained autoencoder containing data domain-specific fine-grained information of in-distribution normal samples.

\vspace{-0.2cm}
\section{Proposed method}
\label{sec:method}
Our method consists of two parts: (a) training an autoencoder to learn useful feature representations, and 
(b) using a Multi-Scale Deep Feature Sparse Coding (MDFSC) to get a residual loss for each test sample. In the following paragraphs, we discuss the proposed method in more details. 

\vspace{-0.2cm}
\subsection{Autoencoder training} 
\begin{figure}[h]
\centering
\includegraphics[width=0.45\textwidth]{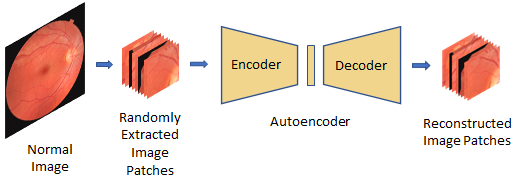}
\vspace{-15px}
\caption{Autoencoder pipeline block diagram. Random patches are extracted from each image in the training set to train the autoencoder.} 
\label{Autoencoder_pipeline}
\end{figure}

We have utilized a VGG-based \cite{vgg} autoencoder in our work. We have used feature extractor of VGG-16 as encoder of our autoencoder. Hence, the encoder consists of 13 convolutional layers and 5 Maxpooling layers. Then, the encoded feature map is flattened and passed to a fully connected layer, producing a latent vector of length 256. This latent vector is passed to another linear layer and the output is reshaped to a 3D tensor. Further, this 3D tensor is fed to decoder. The decoder is constructed symmetrically with the encoder. We have trained this autoencoder network on the normal samples of the training subset. Figure-\ref{Autoencoder_pipeline} shows block diagram for autoencoder pipeline.

\vspace{-0.2cm}
\subsection{Multi-Scale Deep Feature Sparse Coding}
The central idea of anomaly detection with sparse coding is to reconstruct a given image using a dictionary learnt exclusively from normal images and then threshold the reconstruction error as a residual error. Sparse Coding facilitates in the extraction of relevant features from multi-scale features. Since the dictionary is trained to represent only normal images, the reconstruction error is likely to be low if the provided input image is normal. The reconstruction error will be large if the input image contains anomalies that the normal dictionary is unable to represent.

In this part, we extract multi-scale features from the encoder part of the trained autoencoder. Specifically, we obtain features from several layers at different depths of the encoder because the intermediate features of the encoder contain micro and macro-level information about the input image. These multi-scale features are used to form the input feature matrix for sparse coding. The motivation behind using the multi-scale feature is to increase the receptive field of the extracted features and detect anomalies at different scales. We extract multi-scale features $f^p_1,f^p_2$ and $f^p_3$ from the encoder part of the trained autoencoder given an input image patch $I_p$ in the following manner. 
\begin{gather*}
    f^p_1 = \phi_{21}(I_p), \qquad f^p_2 = \phi_{14}(T_d(I_p,2)) \\
    f^p_3 = \phi_{7}(T_d(I_p,4)) 
\end{gather*}
where $\phi_L(x)$ extracts feature activation of $L$-th layer given the input $x$, and $T_d(x,s)$ downscales input $x$ by a factor of $s$.

\begin{figure*}[h]
\centering
\includegraphics[width=0.8\textwidth]{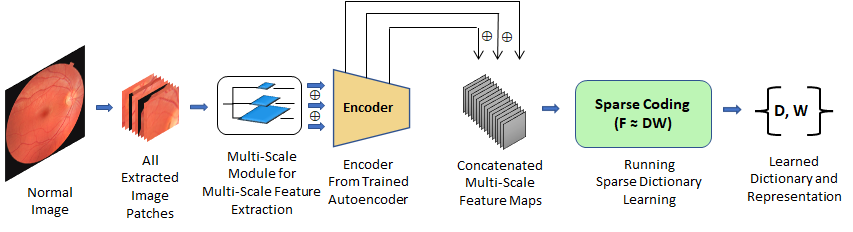}
\vspace{-15px}
\caption{MDFSC training pipeline block diagram. First, image patches are extracted from input images. Each patch is then appropriately rescaled and fed to trained encoder part of autoencoder to generate multi-scale features. Then, these features are concatenated to form input feature matrix for sparse coding that learns a dictionary and sparse representation.} 
\label{MDFSC inference}
\vspace{-15px}
\end{figure*}

Similar to Sparse coding \cite{imamura2021mlf}, MDFSC also decomposes input multi-scale feature tensors, $\mathbf{F}= \left[\mathbf{f}_{1}, \ldots, \mathbf{f}_{m}\right], \mathbf{f}_{i} \in \mathbb{R}^{d}$ as a linear combination of a few basis vectors, $\mathbf{d}_{i}$ $(i=1, \ldots, n)$ from a dictionary matrix, $\mathbf{D}=\left[\mathbf{d}_{1}, \ldots, \mathbf{d}_{n}\right], \mathbf{D} \in \mathbb{R}^{d \times n}$ and sparse coefficients, $\mathbf{w}_{i}$ $(i=1, \ldots, m)$ from sparse representation, $\mathbf{W}=\left[\mathbf{w}_{1}, \ldots, \mathbf{w}_{m}\right], \mathbf{w}_{i} \in \mathbb{R}^{n}$. The dictionary used for sparse representation is learned by minimizing the following objective function.
$$
\min _{\mathbf{D} \in \mathcal{C}, \left(\mathbf{w}_{i}\right)_{i=1}^{m}} \sum_{i=1}^{m}\left(\frac{1}{2}\left\|\mathbf{f}_{i}-\mathbf{D} \mathbf{w}_{i}\right\|_{2}^{2}+\alpha\left\|\mathbf{w}_{i}\right\|_{1}\right)
$$
where $\mathcal{C} \equiv\left\{\mathbf{D} \in \mathbb{R}^{d \times n}:\left\|d_{i}\right\|_{2} = 1, \forall i=1, \ldots, n\right\}$ and $\alpha>0$ is a regularization co-efficient. Training pipeline for MDFSC is shown in Figure-\ref{MDFSC inference}.

\vspace{-0.2cm}
\subsection{Inference}
We first extract all the image patches from the input image during inference. These patches are resized and passed to encoder to obtain multi-scale features. Extracted multi-scale features are concatenated to form a feature matrix. We obtain sparse representation and residual loss using learnt dictionary.

We define an input image's anomaly score as the sum of the top-k highest residual losses as reconstruction errors from all input multi-scale feature patches. As a result, minor error zones are discarded, leaving just potentially abnormal regions. This is due to the fact that even an abnormal image contains numerous normal areas. An anomalous image with a minor anomaly will be incorrectly categorized as normal if the reconstruction errors of all the multi-scale feature patches are added together.
\vspace{-15px}
\section{Experiments}
\label{sec:exp}
\vspace{-0.2cm}
\subsection{Dataset description}
\vspace{-0.2cm}
This paper has used IDRiD, Eye-Q and OCTID datasets for training and evaluation.

\textbf{Eye-Q:} Eye-Quality (Eye-Q) \cite{fu2019evaluation} dataset contains 28,792 retinal images divided into three categories: ``Good", ``Usable" and ``Reject". In this work, we have considered ``Good" and ``Usable" categories as normal and ``Reject" category as anomalous. There are 23,252 normal and 5,540 anomalous images in this dataset.

\textbf{IDRiD:} Indian Diabetic Retinopathy Image Dataset (IDRiD) \cite{porwal2018indian} is a fundus image dataset consisting of 516 images. These images belong to two categories: Normal images and Retinal images with signs of Diabetic Retinopathy and Diabetic Macular Edema. We treat the later category as anomalous images in our experiments. There are 348 normal and 168 anomalous samples in this dataset.

\textbf{OCTID:} Optical Coherence Tomography Image Database (OCTID) \cite{gholami2020octid} contains 470 high resolution OCT images. These images are categorized into normal and four disease categories: Macular Hole, Age-related Macular Degeneration, Central Serous Retinopathy, and Diabetic Retinopathy. We have considered the disease categories as anomalous samples. In this dataset, there are 264 normal and 206 anomalous images. 

\vspace{-0.2cm}
\subsection{Training and Evaluation details} 

We have used a machine with Intel Xeon CPU with 32 GB RAM and NVIDIA 1080Ti GPU card with approximately 12 GB GPU Memory for all of our experiments. 

\textbf{Training of autoencoder:} We resize the input images to $512\times512$ and randomly crop patch of size $64\times64$ during training. We have used $L_2$ loss to train the network. We used Adam optimizer \cite{kingma2014adam} with initial learning rate set to $10^{-4}$. 

\textbf{Training of MDFSC:} 
We have trained the MDFSC module with normal images only from the training data by constructing a learned dictionary matrix for each type of dataset. we have resized and normalized our input normal image to $512 \times 512$ as preprocessing. After that, image patches of $16 \times 16$ size are extracted with stride of 2 from the preprocessed image. Multi-scale feature patches were extracted from
$conv2\_2, conv3\_3, conv4\_3$
layers of the VGG16 encoder sub-network of the autoencoder with respect to the decreasing scale of the image patches. We have used 50 basis vectors for constructing dictionary matrix and Lasso-LARS \cite{efron2004least} is used for the loss optimization. The value of $\alpha$ is chosen to be 1 in our experiments empirically.

\textbf{Evaluation metrics.} We have used Average Precision (AP) and Area Under Curve (AUC) scores to evaluate the performance of different models. 
Average Precision\footnote{\url{https://scikit-learn.org/stable/modules/generated/sklearn.metrics.average_precision_score.html}} is the weighted mean of precisions achieved at each threshold, with the increase in recall from the previous threshold used as the weight.
Area Under Curve score is the area under  Receiver Operating Characteristic (ROC) curve. 
\vspace{-0.2cm}
\subsection{Comparison with state-of-the-art}
We have compared our approach with multiple state-of-the-art methods: AnoGAN \cite{anogan}, Deep-Geo \cite{deep_geo}, Deep-IF \cite{deep_if} and SPADE \cite{SPADE}. The quantitative results are shown in Table-\ref{sota_table}. Our method outperforms other state-of-the-art methods on all the datasets. Deep-IF \cite{deep_if} and SPADE \cite{SPADE} are two closest methods to our method as they also use aggregated deep features from large Imagenet pretrained networks, but they also suffer from data domain related issue as medical images are quite different from images from Imagenet. 

\begin{table}
\small
\centering
\label{sota_table}
\begin{tabular}{|c|c|c|c|c|} 
\hline
\multirow{2}{*}{\textbf{Methods}}                                    & \multirow{2}{*}{\textbf{Metric}} & \multicolumn{3}{c|}{\textbf{Dataset}}             \\ 
\cline{3-5}
                                                                     &                                  & \textbf{Eye-Q} & \textbf{IDRiD} & \textbf{OCTID}  \\ 
\hline
\multirow{2}{*}{AnoGAN \cite{anogan}}                                              & AP                               & 0.7763         & 0.5264         & 0.7183          \\ 
\cline{2-5}
                                                                     & AUC                              & 0.7036         & 0.5068         & 0.6568          \\ 
\hline
\multirow{2}{*}{Deep-Geo \cite{deep_geo}}                                            & AP                               & 0.8183         & 0.5857         & 0.7466          \\ 
\cline{2-5}
                                                                     & AUC                              & 0.7897         & 0.5580         & 0.7852          \\ 
\hline
\multirow{2}{*}{Deep-IF \cite{deep_if}}                                             & AP                               & 0.8410         & 0.6255         & 0.7966          \\ 
\cline{2-5}
                                                                     & AUC                              & 0.8221         & 0.6600         & 0.8186          \\ 
\hline
\multirow{2}{*}{SPADE \cite{SPADE}}                                               & AP                               & 0.8656         & 0.6212         & 0.7645          \\ 
\cline{2-5}
                                                                     & AUC                              & 0.8232         & 0.6556         & 0.8240          \\ 
\hline
\multirow{2}{*}{\begin{tabular}[c]{@{}c@{}}Our\\method\end{tabular}} & AP                               & \textbf{0.9098}         & \textbf{0.6576}         & \textbf{0.9418}          \\ 
\cline{2-5}
                                                                     & AUC                              & \textbf{0.8874}         & \textbf{0.6997}         & \textbf{0.9243}          \\
\hline
\end{tabular}
\caption{Quantitative comparison with state-of-the-art methods on Eye-Q, IDRiD and OCTID datasets.}
\vspace{-12px}
\end{table}
\subsection{Ablation Study}

\textbf{Importance of training on medical datasets.} 
We have compared the performance of features extracted from our trained encoder and Imagenet-trained VGG16 network \cite{imamura2021mlf}. We can observe in Table-\ref{abl1} that features from trained autoencoder achieves higher AUC score for both the datasets. This can be attributed to the fact that when the network is trained on medical image datasets, it can learn features that are more relevant to medical data than features extracted from an Imagenet-trained network.
\begin{table}[h]
\small
\centering
\begin{tabular}{|c|c|c|c|} 
\hline
\textbf{Dataset}       & \textbf{Metric} & \begin{tabular}[c]{@{}c@{}}\textbf{Our }\\\textbf{method}\end{tabular} & \begin{tabular}[c]{@{}c@{}}\textbf{\textbf{Using}}\\\textbf{\textbf{Imagenet}}\\\textbf{\textbf{features}}\end{tabular}  \\ 
\hline
\multirow{2}{*}{EyeQ}  & AP              & \textbf{0.9098}                                                                 & 0.7523                                                                                                                           \\ 
\cline{2-4}
                       & AUC             & \textbf{0.8874}                                                                 & 0.6840                                                                                                                           \\ 
\hline
\multirow{2}{*}{IDRiD} & AP              & \textbf{0.6446}                                                                 & 0.6235                                                                                                                           \\ 
\cline{2-4}
                       & AUC             & \textbf{0.6897}                                                                 & 0.5957                                                                                                                           \\ 
\hline
\multirow{2}{*}{OCTID} & AP              & \textbf{0.9418}                                                                 & 0.8463                                                                                                                           \\ 
\cline{2-4}
                       & AUC             & \textbf{0.9243 }                                                                & 0.7866                                                                                                                           \\
\hline
\end{tabular}
\caption{Ablation study of classification performance: Features learnt from medical data vs. Imagenet features.}
\label{abl1}
\end{table}

\vspace{-0.2cm}
\textbf{Comparison with anomaly detection using Reconstruction loss.} We have trained an autoencoder with linear layers removed to directly compute reconstruction error (MSE) of test images. This reconstruction error is used as anomaly score \cite{zhou2017anomaly} to calculate AP and AUC scores. In Table-\ref{abl2}, we have compared the performance of anomaly detection using reconstruction error with our approach. We can see that our approach is superior to only autoencoder reconstruction loss based approach. With only autoencoder, it learns fine-grained information of normal samples and a few other non-discriminative redundant features due to memorization. With additional sparse coding, it extracts multi-scale features which are common across the samples in the dataset for anomaly detection. 

\begin{table}[h]
\small
\centering
\begin{tabular}{|c|c|c|c|} 
\hline
\textbf{Dataset}       & \textbf{Metric} & \begin{tabular}[c]{@{}c@{}}\textbf{Our }\\\textbf{Method}\end{tabular} & \begin{tabular}[c]{@{}c@{}}\textbf{Autoencoder}\\\textbf{Recons. loss}\\\textbf{\textbf{based method}}\end{tabular}  \\ 
\hline
\multirow{2}{*}{EyeQ}  & AP              & \textbf{0.9098}                                                                 & 0.6874                                                                                                  \\ 
\cline{2-4}
                       & AUC             & \textbf{0.8874}                                                                 & 0.8014                                                                                                  \\ 
\hline
\multirow{2}{*}{IDRiD} & AP              & \textbf{0.6446}                                                                 & 0.5107                                                                                                  \\ 
\cline{2-4}
                       & AUC             & \textbf{0.6897}                                                                 & 0.5414                                                                                                  \\ 
\hline
\multirow{2}{*}{OCTID} & AP              & \textbf{0.9418}                                                                 & 0.8815                                                                                                  \\ 
\cline{2-4}
                       & AUC             & \textbf{0.9243}                                                                 & 0.8964                                                                                                  \\
\hline
\end{tabular}
\caption{Ablation study of Classification Performance with Autoencoder Reconstruction loss based method}
\label{abl2}
\end{table}

\vspace{-0.2cm}
\subsection{Qualitative results}
\begin{figure}[h]
\centering
\includegraphics[scale=0.95]{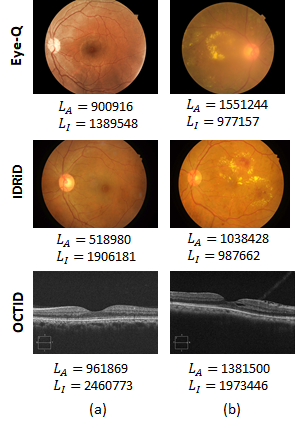} 
\vspace{-15px}
\caption{Qualitative results. (a) Normal samples, (b) Anomalous samples. Here, $L_A$ stands for anomaly score computed by our approach and $L_I$ denotes anomaly score calculated using Imagenet features. The lower the anomaly score, the better.
} 
\vspace{-15px}
\label{Visual results}
\end{figure}

In Figure-\ref{Visual results}, we have compared anomaly scores computed using features from our trained autoencoder with anomaly scores computed using Imagenet-trained features. We can see that anomaly score for our approach is lower for normal samples. On the other hand, anomaly score using Imagenet features are higher in case of anamalous samples in our approach.

\section{Conclusion}
In this paper, we have presented an unsupervised approach for detecting anomalies in medical images. We utilize multi-scale features extracted from a trained autoencoder to learn dictionary in sparse coding. At evaluation time, we calculate reconstruction error for different patches and sum top 5 reconstruction errors as anomaly score. In ablation study, it has been shown that autoencoder training step helps to detect out-of-distribution pixels from the anomaly images and boosted the performance of the MDFSC anomaly detection pipeline by a large margin. Our method surpasses multiple state-of-the-art methods in three datasets covering most of the common ocular diseases. In future, vision transformer \cite{vit} based encoders can be explored to extract more enriched features from the image and evaluate this idea for more fine-grained classification tasks.

\bibliographystyle{IEEEbib}
\bibliography{strings,refs}

\end{document}